# Securing Biometric Images Using Reversible Watermarking

**Sabu M. Thampi**
Rajagiri School of Engineering and Technology, Kochi, India

sabum@rajagiritech.ac.in

**Ann Jisma Jacob**
Indira Gandhi National Open University, India

annjisma@yahoo.com

## Abstract

Biometric security is a fast growing area. Protecting biometric data is very important since it can be misused by attackers. In order to increase security of biometric data there are different methods in which watermarking is widely accepted. A more acceptable, new important development in this area is reversible watermarking in which the original image can be completely restored and the watermark can be retrieved. But reversible watermarking in biometrics is an understudied area. Reversible watermarking maintains high quality of biometric data. This paper proposes Rotational Replacement of LSB as a reversible watermarking scheme for biometric images. PSNR is the regular method used for quality measurement of biometric data. In this paper we also show that SSIM Index is a better alternate for effective quality assessment for reversible watermarked biometric data by comparing with the well known reversible watermarking scheme using Difference Expansion.

## 1. Introduction

Applying security along with privacy is challenging in this digital age. Conventional methods are defeated by biometric authentication as it is more convenient for privacy protection. But along with convenience it has introduced new threats. It is prone to attacks since the data will not be renewed so often. Thus securing biometric data has become crucial. Though there are different methods being introduced for this, watermarking plays an important role here in protecting data.

Watermarking hides a message or image in some data to obtain a new data so that the hidden data is indistinguishable from the original data so that an attacker cannot remove or replace the message from the new data. Thus it has become a major part of information hiding. Watermarking is used to pass hidden messages in communications. Watermarking can be done in different domains like spatial or frequency domain. In Spatial domain actual pixel values are changed where as in the frequency domain, coefficients used in transformed image representations are changed. There are some techniques, which uses LSBs for embedding watermark data. In most of these, the changes made to the original data become irreversible. There is yet another way in which the original image and the embedded payload can be perfectly recovered later by watermarking in some particular way. This technique is called *reversible watermarking.*

In Reversible watermarking, a watermark is embedded into an image and from the watermarked image; original image and the watermark are retrieved. The retrieval process does not need the original image and so this technique is said to be blind [7]. There are different types of reversible watermarking based on the technique they use such as Data Compression, Difference expansion, Histogram operation and LSB Replacement [8]. Figure 1 shows the difference between the conventional watermarking techniques and the reversible watermarking techniques. This paper focuses on reversible watermarking using difference expansion and LSB replacement.

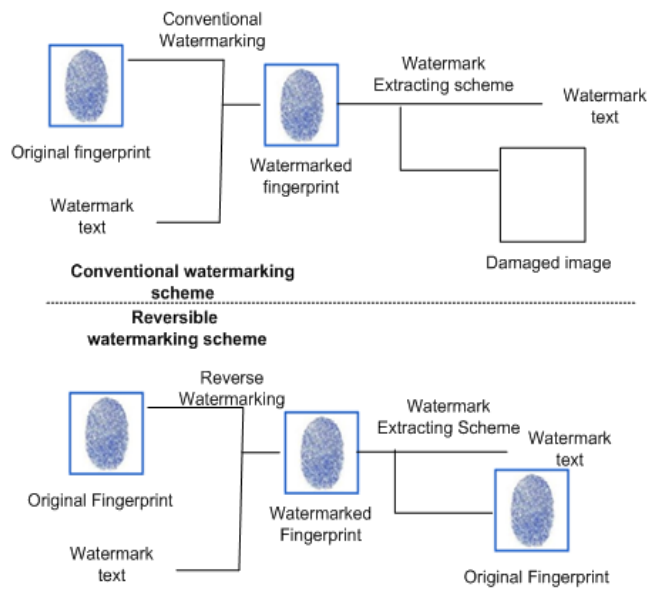

**FIGURE 1:** Difference between conventional and Reversible watermarking system

Quality of watermarked images plays an important role in biometrics. Since watermarking adds noise to the data, it should be done in such a way that it will not defeat the purpose of watermarking. There are quite a few number of techniques introduced for watermarking but the quality of watermarking is usually measured using PSNR. PSNR is a simple measure of image quality based on average error. This technique is known to be not very accurate [5], [6]. Despite its drawbacks, however, PSNR is still heavily reported because it is easy to compute. This paper introduces SSIM Index as a better way of quality measurement for biometric watermarked images. For this, the proposed rotational replacement of LSB scheme is being compared with the well known reversible watermarking scheme using Difference Expansion.

## 2. PREVIOUS WORK

Earliest Reversible watermarking scheme was invented by Barton [3]. DE is the most established method for reverse watermarking [2]. Reversible Watermarking using DE was first proposed by Tian [2]. Tian used regular images for his experiments. This paper uses DE method for watermarking biometric image.

PSNR is the common quality measurement technique used for watermarked images as it is easy to calculate and optimize. But they are not very well matched to perceived visual quality [1]. SSIM (Structural SIMilarity) Index is based on Full-reference image quality assessment [1]. SSIM Index compares local patterns of pixel intensities that have been normalized for luminance and contrast [1]. Since SSIM Index is designed to improve on traditional methods like PSNR, measuring the quality using SSIM Index will add value to the reversible watermarking technique.

In this paper, reversible watermarking is done using proposed rotational replacement of LSB and also using Tian's reversible watermarking technique. The quality of watermarked images is assessed using SSIMIndex and compared.

## 3. WATERMARKING USING DIFFERENCE EXPANSION

Tian [2] introduced the Difference Expansion technique. In this technique, pairs of pixels of the host image $HI$ are identified and they are converted to a low-pass image $LI$ using integer transformation with integer average $\alpha$ and a high-pass image $HI$ containing the pixel differences $\delta$. If x and y be the intensity values of a pixel-pair, then α and δ are defined as

$$\alpha = \left\lfloor \frac{(x+y)}{2} \right\rfloor \qquad\qquad (1)$$

$$\delta = x - y \qquad\qquad\qquad (2)$$

x and y can be computed from α and δ as

$$x = \alpha + \left\lfloor \frac{(\delta+1)}{2} \right\rfloor \qquad\qquad (3)$$

$$y = \alpha - \frac{\delta}{2} \qquad\qquad (4)$$

By appending an information bit i to the LSB of the difference δ, a new LSB can be created. The watermarked difference is

$$\delta_w = 2\delta + i \qquad\qquad (5)$$

The resulting pixel gray-levels are calculated from the difference ($\delta_w$) and integer average $\alpha$ using (3) and (4).

For an image with n-bit pixel representation, the gray levels satisfy $x, y \in [0, 2^n - 1]$, if and only if α and δ satisfies the following condition:

$$|\delta| \in R_d(\alpha) = [0, \min(2(2^n - 1 - l), 2\alpha + 1)] \qquad (6)$$

Where $R_d(\alpha)$ is called the invertible region. Combining (5) and (6) we obtain the condition for a difference δ to undergo DE.

$$|2\delta + i| \in R_d(\alpha) \text{ for i = 0,1} \qquad\qquad (7)$$

This condition is called the *expandability condition* for DE. If an integer average is given and the difference satisfies the expandability condition, it is called an *expandable difference*.

Another method used by Tian in his Difference Expansion technique, other than embedding which is already discussed is replacing LSB. Here, an information bit will be used to replace the LSB of the difference. Since the LSB is replaced in the embedding process, this cannot be considered as lossless as the first embedding technique. However, here the information about the true LSBs of the differences that are embedded by replacing LSB are saved and embedded with the payload, to ensure that it is not lossy.

The LSB of a difference can be flipped without affecting its ability to invert back to the pixel domain if and only if

$$\left| 2\left\lfloor \frac{\delta}{2} \right\rfloor + i \right| \in R_d(\alpha) \text{ for i=0, 1} \qquad\qquad (8)$$

This is called the *changeability condition*. A difference satisfying the changeability condition, given a corresponding integer average, is called a *changeable difference*. An expandable difference is also a changeable difference. A changeable location remains changeable even after its LSB is replaced, whereas an expandable location may not be expandable after DE, but it remains changeable.

## 4. WATERMARKING USING ROTATIONAL REPLACEMENT OF LSB

In common LSB replacement techniques, the LSB of the original image is replaced with the MSB of the watermark. For this, first the number of bits needs to be required is calculated and those many LSB are replaced with MSB of watermark. To retrieve the watermark, first get the LSB according to the number used and then use them to create new image by changing them to MSB. Here, the restored image will be low in quality. In Rotational replacement of LSB, same bit will be used to store secondary LSB of the original image and the watermark.

For embedding watermark using rotational replacement of LSB, first create matrices of original image and watermark. Assuming original image is X × Y matrix and watermark is x×y matrix. The ratio γ = (M×N)/m×n should be greater than or equal to 8, so that watermark can be embedded using this technique. The original image matrix is divided into blocks. Thus if γ ≥64, the block size will be 8×8. Only one byte of watermark data can be embedded to such a block. Consider such a block. The last row of the original host image will be replaced by the last but one row data. This will be continued rotationally so that the first row of the original matrix will be placed at the second row and first row will be free. One byte of the watermark data will be placed in this row.

$$\begin{bmatrix} 11111111 \\ 10101010 \\ 01010101 \\ 11011011 \\ 00110100 \\ 11100010 \\ 10001000 \\ 11000001 \end{bmatrix} + \begin{bmatrix} 10101010 \end{bmatrix} \rightarrow \begin{bmatrix} 10101010 \\ 11111111 \\ 10101010 \\ 01010101 \\ 11011011 \\ 00110100 \\ 11100010 \\ 10001000 \end{bmatrix}$$

**FIGURE 2:** Rotational replacement of LSB while embedding

Extraction will be similar to embedding process where, using the γ value, blocks are determined. Based on the size of the watermarked data, first row LSB will be used to reconstruct the watermark. After removing the first row, the remaining rows are rotationally replaced in the reverse order of embedding. Last row will replace the data of the same row.

## 5. STRUCTURAL SIMILARITY INDEX (SSIM INDEX)

SSIM Index is a framework for quality assessment based on the degradation of structural information introduced by Zhou Wang [1] *et al.* For human visual system a calculation of structural information difference can provide a good approximation to the image distortion perceived. The product of the illumination and the reflectance gives the luminance of the surface of an object, but the structures of the objects in the scene are independent of the illumination [1]. SSIM Index defines the structural information in an image as those attributes that represent the structure of objects in the scene, independent of the average luminance and contrast [1].

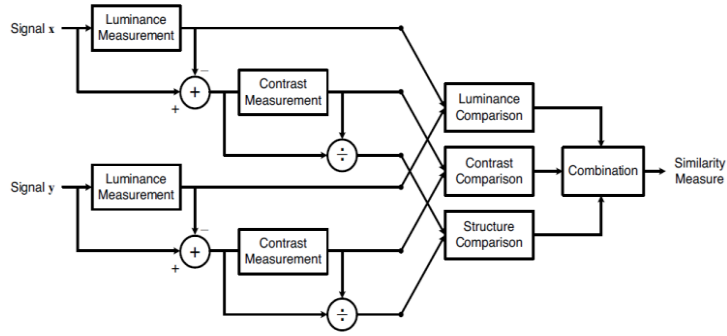

**FIGURE 3:** Diagram showing calculation of SSIM Index [1]

In SSIM Index, the similarity measurement is done by making comparison of luminance, contrast and structure. Let x and y be two image signals.

Luminance of each signal is calculated as:

$$\mu_x = \frac{1}{N}\sum_{i=1}^{n} x_i \tag{1}$$

Signal contrast is calculated as:

$$\sigma_x = (\frac{1}{N-1}\sum_{i=1}^{n}(x_i - \mu_x)^2)^{\frac{1}{2}} \tag{2}$$

The structure comparison is calculated as:

$$(x - \mu_x)/\sigma_x \text{ and } (y - \mu_y)/\sigma_y \tag{3}$$

Combining these three, the SSIM index is calculated as:

$$\text{SSIM}(x,y) = \frac{(2\mu_x\mu_y + C_1)(2\sigma_{xy} + C_2)}{(\mu_x^2\mu_y^2 + C_1)(\sigma_x^2 + \sigma_y^2 + C_2)}$$

Where $C_1$ & $C_2$ are included to avoid instability when $\mu_x^2 + \mu_y^2$ is very close to zero.

$C_1 = (K_1 L)^2$ where $K_1 = 0.01$
$C_2 = (K_2 L)^2$ where $K_2 = 0.03$
L = dynamic range of the pixel values.

## 6. DESIGN OVERVIEW

The novelty of this paper is in introducing Rotational Replacement of LSB as a reversible watermarking scheme and proving SSIM Index as a better alternate for measuring quality of biometric data. Since PSNR is easy to calculate, it is the common method used everywhere, where as the experiments conducted show that SSIM Index can be used as a better and effective alternate. The system uses Biometric image (fingerprint) for watermarking using Rotational Replacement of LSB and also using Difference Expansion. These reversible watermarking techniques reduce the noise while watermarking which is highly recommended for biometric images. The watermarked fingerprint image is used for calculating SSIM index. SSIM Index is calculated for watermarked images with different payload.

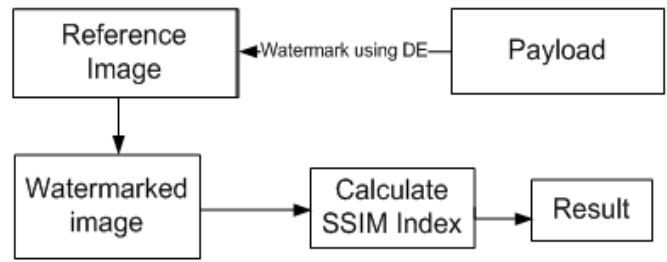

**FIGURE 4:** Calculation of SSIM Index of watermarked image

## 7. EXPERIMENTAL SETUP AND RESULTS

The experiment was conducted using a single fingerprint image with different payloads since the quality of image is affected more as the payload increases. The finger print was collected from internet [http://www.nist.gov] and it is of the resolution 512*512. The experiment was implemented in Java using Image APIs. Watermarking was done using Rotational Replacement of LSB and DE for different payload. SSIM index for the watermarked images are found. The goal is to prove that the SSIM Index based quality measurement is a better alternate than PSNR for biometric image watermarking and also that Rotational replacement of LSB is a better reverse watermarking scheme than DE.

Fig 5 shows the original image and the watermarked images used for the experiment using rotational replacement of LSB. As it can be seen, the images are visually not tampered by watermarking.

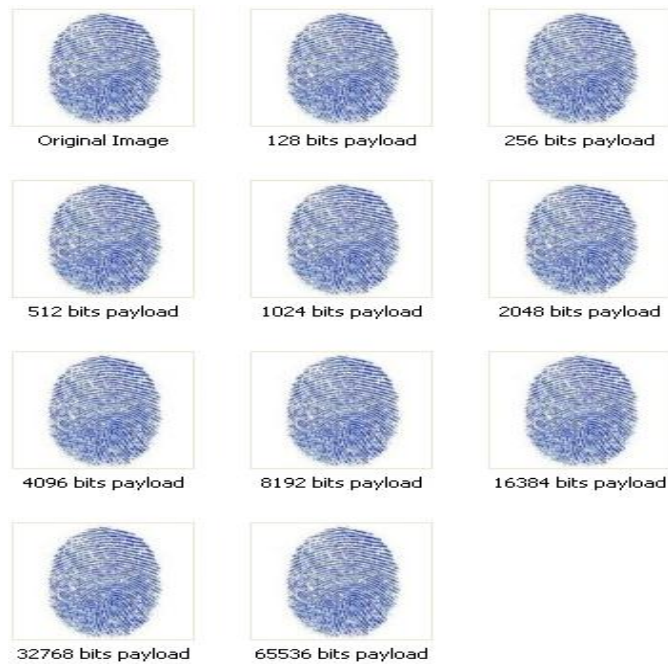

**FIGURE 5:** Watermarked samples with different payload

Experimental results show that Rotational replacement of LSB is a better reverse watermarking scheme than DE by measuring the quality using SSIMIndex. It is also proved that the execution time for Rotational replacement of LSB is much less than DE and payload capacity of Rotational replacement of LSB is more than DE The SSIM Index for all the watermarked images mentioned here were found out. Figure 6 shows that SSIMIndex for DE watermarked images has a lower value indicating more noise and Figure 7 shows that the Rotation LSB replacement is faster than DE for a given payload.

| Payload | DE | LSB |
|---------|-----|-----|
| 128 | 1.00000000 | 1.00000000 |
| 256 | 1.00000000 | 1.00000000 |
| 512 | 1.00000000 | 1.00000000 |
| 1024 | 1.00000000 | 1.00000000 |
| 2048 | 1.00000000 | 1.00000000 |
| 4096 | 1.00000000 | 1.00000000 |
| 8192 | 1.00000000 | 1.00000000 |
| 16384 | 1.00000000 | 1.00000000 |
| 32768 | 0.99999990 | 1.00000000 |
| 65536 | 0.99999936 | 1.00000000 |

**TABLE 1:** Value of SSIMIndex for different payloads using DE and RRL

| Payload | DE | LSB |
|---------|------|-----|
| 128 | 3954 | 688 |
| 256 | 3937 | 687 |
| 512 | 3953 | 719 |
| 1024 | 3968 | 672 |
| 2048 | 4125 | 703 |
| 4096 | 3984 | 672 |
| 8192 | 3985 | 703 |
| 16384 | 4500 | 687 |
| 32768 | 4953 | 703 |
| 65536 | 6906 | 672 |

**TABLE 2:** Value of execution time for different payloads using DE and RRL

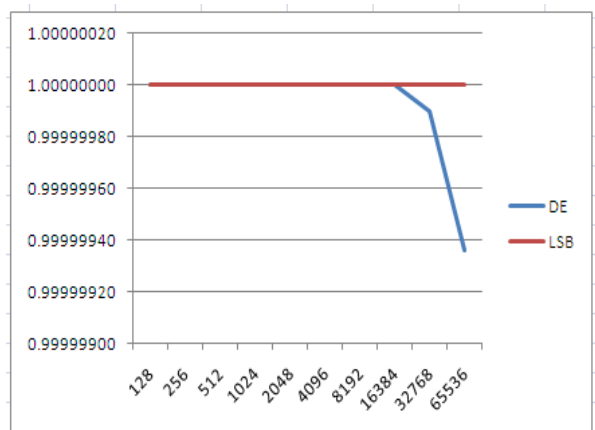

**FIGURE 6:** Graph corresponding to Table 1

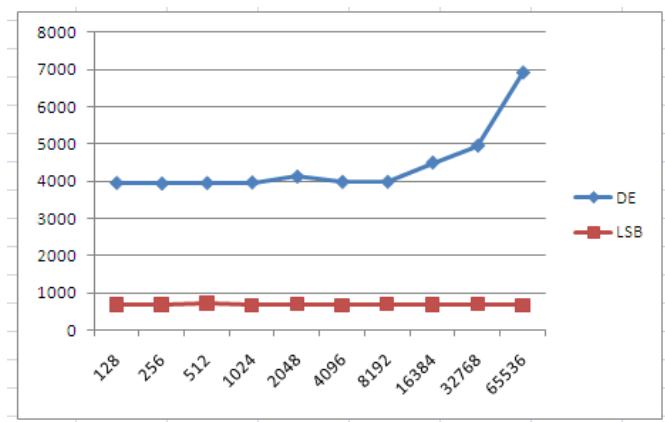

**FIGURE 7:** Graph corresponding to Table 2

## 8. CONCLUSIONS AND FUTURE WORK

Biometric images can be secured using watermarking. Reversible watermarking helps us to have images with minimum distortion. At the same time, the original image can be retrieved for any further use. Rotational Replacement of LSB is a high performing scheme for moderate payload. SSIM Index is proved to be more accurate for measuring the quality of watermarked image. Consequently, in the biometric image security field SSIM Index can be used instead of PSNR to get better result. For reversible watermarking, there is a limitation of maximum payload, depending on the capacity of original image being used. Our future work will be focussing on calculating the performance of feature level extraction for a watermarked image using rotation replacement of LSB.